\documentclass[10pt,twocolumn,journal,subeqn]{IEEEtran}
\usepackage[final]{graphicx}
\usepackage{epsfig}
\usepackage{epstopdf}
\usepackage{latexsym}
\usepackage{amsfonts}
\usepackage{here}
\usepackage{caption}
\usepackage{subcaption}
\usepackage{rawfonts}
\usepackage[T1]{fontenc}
\usepackage{calc}
\usepackage{url}
\usepackage{enumerate}
\usepackage{color}
\usepackage[tbtags]{amsmath}
\usepackage{amssymb}
\usepackage{upref}
\usepackage{times}
\usepackage{amsmath}
\usepackage{cite}
\usepackage{algorithm}
\usepackage{cuted}
\usepackage{multirow}
\usepackage[labelformat=simple]{subcaption}

\newcommand{\bi}{\begin{itemize}}
\newcommand{\ei}{\end{itemize}}

\newcommand{\be}{\begin{enumerate}}
\newcommand{\ee}{\end{enumerate}}
\newcommand{\bd}{\begin{description}}
\newcommand{\ed}{\end{description}}
\newcommand{\bc}{\begin{center}}
\newcommand{\ec}{\end{center}}
\newcommand{\bt}{\begin{tabbing}}
\newcommand{\et}{\end{tabbing}}
\newcommand{\bfig}{\begin{figure}}
\newcommand{\efig}{\end{figure}}
\newcommand{\beq}{\begin{equation}}
\newcommand{\beqarr}{\begin{eqnarray}}
\newcommand{\beqarrn}{\begin{eqnarray*}}
\newcommand{\eeq}{\end{equation}}
\newcommand{\eeqarr}{\end{eqnarray}}
\newcommand{\eeqarrn}{\end{eqnarray*}}
\newcommand{\bflr}{\begin{flushright}\vspace{-0.2in}}
\newcommand{\eflr}{\end{flushright}}
\newcommand{\bsub}{\begin{subequations}}
\newcommand{\esub}{\end{subequations}}
\newcommand{\barr}{\begin{array}}
\newcommand{\earr}{\end{array}}

\begin{document}

\title{Effect of Fog Particle Size Distribution on 3D Object Detection
Under Adverse Weather Conditions}
\author{
\normalsize Ajinkya Shinde,
Gaurav Sharma,~\IEEEmembership{Member,~IEEE},
Manisha Pattanaik,~\IEEEmembership{Senior Member,~IEEE},
Sri Niwas Singh,~\IEEEmembership{Fellow,~IEEE}.
\thanks{Ajinkya Shinde is with the Department of Information Technology Engineering, Atal Bihari Vajpayee-Indian Institute of Information Technology and Management, Gwalior, Madha Pradesh 474015, India (e-mail: {\tt imt\_2019008@iiitm.ac.in.}}
\thanks{Gaurav Sharma, Manisha Pattanaik, and Sri Niwas Singh are with the Department of Electrical and Electronics Engineering, Atal Bihari Vajpayee-Indian Institute of Information Technology and Management, Gwalior, Madha Pradesh 474015, India (e-mail: {\tt sharmagaurav2201@gmail.com, manishapattanaik@iiitm.ac.in, snsingh@iiitm.ac.in).}}\vspace{-2.5em}}%
\maketitle
\begin{abstract}
LiDAR-based sensors employing optical spectrum signals play a vital role in providing significant information about the target objects in autonomous driving vehicle systems. 
However, the presence of fog in the atmosphere severely degrades the overall system's performance. 
This manuscript analyzes the role of fog particle size distributions in 3D object detection under adverse weather conditions. 
We utilise Mie theory and meteorological optical range (MOR) to calculate the attenuation and backscattering coefficient values for point cloud generation and analyze the overall system's accuracy in Car, Cyclist, and Pedestrian case scenarios under easy, medium and hard detection difficulties. 
Gamma and Junge (Power-Law) distributions are employed to mathematically model the fog particle size distribution under strong and moderate advection fog environments. 
Subsequently, we modified the KITTI dataset based on the backscattering coefficient values and trained it on the PV-RCNN++ deep neural network model for Car, Cyclist, and Pedestrian cases under different detection difficulties. 
The result analysis shows a significant variation in the system's accuracy concerning the changes in target object dimensionality, the nature of the fog environment and increasing detection difficulties, with the Car exhibiting the highest accuracy of around 99\% and the Pedestrian showing the lowest accuracy of around 73\%. 

\end{abstract}


\section{Introduction}

Light Detection and Ranging (LiDAR) utilizes the optical spectrum for remote sensing and three-dimensional (3D) environment modeling. Due to the utilization of the optical spectrum, the LiDAR-based systems find their application in oceanography, atmospheric sciences, and monitoring changes in land use patterns \cite{wang2021challenges,debnath2023applications}. 
Recently, LiDAR systems have served as an integral part of autonomous driving vehicles as they create detailed 3D maps of the terrestrial environment, facilitating controlled vehicular navigation in adverse environmental conditions. For surroundings information acquisition, the LiDAR systems employ several sensors that enable the detection of various objects, particularly vehicles, pedestrians, and other moving objects \cite{you2020lidar}. 

Recent developments in LiDAR technology include advancements in miniaturization and functionality to meet the growing demands of the automotive market \cite{9455394}. Integration with multi-sensor fusion simultaneous localization and mapping (SLAM) systems enhances navigation accuracy, particularly in challenging environments \cite{rs14122835}. Airborne LiDAR technology benefits from lightweight sensors, and recent developments in lightweight sensors have made the applications such as power surveying and mapping much more effective \cite{10.1007/978-981-99-6956-2_12}. Mobile LiDAR systems, incorporating LiDAR sensors, inertial measurement units (IMUs), and global positioning system (GPS), offer dynamic 3D mapping capabilities for various sectors, including transportation, urban planning, and robotics \cite{geomatics3040030}. These advancements collectively drive the expansion of LiDAR applications, enhancing mapping accuracy and efficiency across industries \cite{rs5094652}.

3D object detection is essential in applications, 
such as autonomous driving, augmented reality, indoor mapping, drones and geographical mapping. Compared to two-dimensional (2D) images, 3D structures provide extra information about the geographical constitution of the structures \cite{Pan_2021_CVPR}. In computer vision, 3D object detection is a critical component, particularly in the case scenarios of real-world environments such as autonomous systems. Primarily, 3D object detection involves identifying and localising objects in 3D dimensional space using various data sources such as LiDAR, RADAR, and red, green, and blue (RGB) imagery. However, due to the sparsity and complexity of the point cloud data (representing the object's 3D coordinates), 3D object detection is challenging. To overcome the challenges, 
recently, the fusion of RGB and point cloud data is proposed \cite{wang2020overview}.

Monocular camera-based 3D object detection, stereo camera-based 3D object detection, and LiDAR-based 3D object detection are the three most popular techniques for generating 3D bounding boxes. These bounding boxes are essential for articulating the dimensional features of the object. Among the aforementioned techniques, the monocular camera-based methodology employed in monocular 3D (Mono3D) \cite{Chen20162147} and GS3D \cite{Li20191019} algorithms utilizes 2D images to create 3D bounding boxes. On the contrary, the stereo-camera-based detection technique employs cameras to measure the depth feature of the image as an additional channel. Commonly in algorithms such as pseudo-LiDAR++ \cite{qian2020end}, disparity-based region-based convolutional neural network (Disp R-CNN) \cite{sun2020disp}, and deep stereo geometry network (DSGN) \cite{chen2020dsgn}, stereo cameras are utilized for 3D object detection. However, monocular and stereo camera-based techniques do not yield accurate results for autonomous vehicles and space applications \cite{amzajerdian2016imaging}. 

LiDAR-based 3D object detection techniques are utilized in space and autonomous vehicle applications for higher accuracy. The data obtained from LiDAR-based systems are then processed through grid-based or point-cloud-based methodologies. In grid-based methods, the point cloud is divided into grids, and subsequently, convolutional neural networks (CNN) are used to process the point clouds. For example, in the multi-view 3D object detection network (MV3D) \cite{chen2017multi}, the point clouds are projected in a 2D birds-eye view (which is divided into grids) for detection. Based on the 2D bird-eye-view object detection, bounding boxes are utilized to extract the characteristic features of the images, which are subsequently used to re-create the 3D space. Compared to the grid-based methods, the point-based methods directly process in point clouds for object detection and generation of 3D bounding boxes. Some examples of point-based methods are PointNet \cite{qi2018frustum}, and STD \cite{yang2019std}.

Recently, a new algorithm point voxel region-based convolution neural network (PV-RCNN) \cite{shi2020pv} that utilizes the advantages of both grid-based and point-based techniques was proposed. PV-RCNN is a high-performance 3D object detection framework that integrates 3D voxel CNN and point-based set abstraction to learn discriminative point cloud features. High-quality 3D proposals are generated by the voxel CNN.
On the contrary, PV-RCNN++ \cite{shi2023pv}, being an advanced version of PV-RCNN, provides a complete framework for 3D object detection. It introduces sectorized proposal-centric sampling and VectorPool aggregation to efficiently produce more representative key points and aggregate local point features with less resource consumption.
In \cite{hahner2021fog}, the authors analyzed various 3D object detection algorithms. and found that PV-RCNN \cite{shi2020pv} works best in adverse foggy weather conditions. Our study utilises PV-RCNN++ for analysis.

In addition to all these features, foggy conditions play a vital role in observing the overall characteristics of autonomous vehicles from a 3D object detection perspective.
Fog adversely affects LiDAR and other sensor information by scattering and absorbing electromagnetic waves, which, in the LiDAR system, are the primary sources of information. 
The absorption and the scattering phenomenon results in visibility reduction and leads to the introduction of data errors \cite{zhan2021impact}. 
Usually, the Mie scattering theory incorporates the impact of scattering through the particle size distribution and is utilized to model scattering effects in the current literature. 
In \cite{maitra1996propagation}, the impact of attenuation and dispersion phenomenon on optical pulses passing through a fog-filled medium were studied. 
The role of heavy fog and smoke on the bit error rate of a high-speed free space laser communication was analyzed in \cite{strickland1999effects}. On the contrary, the presence of aerosols in the atmosphere, and its impact on the filament patterns of TW laser beams was studied in \cite{mejean2005multifilamentation}. In addition, numerous studies \cite{Kedar:03,awan2009characterization,fischer2008optical,4656761,5379065,6679219,rashed2015optical,sree2020estimation} analyzed the role of fog, smoke, and other precipitation on free-space optical links. 

In the existing literature, the size distribution is usually modeled through log-normal distributions where the diameter of the fog droplets ranges from a few to tens of micrometres ($\mu$m). 
For excellent sensor performance under foggy conditions, it is imperative to model the particle size distribution of the fog droplets accurately \cite{s23156891}. 
Currently, due to the lack of availability of LiDAR data in foggy conditions, \cite{hahner2021fog} suggested a way to simulate fog particles in already collected LiDAR data by evaluating attenuation and backscattering coefficient values based on the Meteorological Optical Range (MOR). 
On the contrary, a fog particle simulation under particle size distribution was proposed \cite{liu2022parallel}. 

Currently, considerable LiDAR data is available under clear environmental conditions. However, the simultaneous existence of fog particles in the environment causes severe attenuation, scattering, and absorption of the optical signals, subsequently deteriorating the system's overall performance \cite{hahner2021fog}. 
Therefore, there is a considerable requirement for foggy datasets on which deep-learning-based models can be trained. The lack of foggy data is due to the high cost and difficulty of collecting LiDAR data in fog conditions. This paper deals with a generalized way of simulating fog particles using various particle size distributions.

Disturbances in the LiDAR sensor data due to rainy and foggy conditions were studied with an experimental setup in \cite{app11073018}. The authors employed the Kalman filter for noise filtering and real-point cloud prediction. Several simulation techniques have been proposed to create artificial fog in images. For instance, \cite{sakaridis2018semantic} created a foggy version of the cityscapes dataset \cite{cordts2016cityscapes}, which is used for semantic segmentation. In addition, a similar version of the Synscapes dataset was generated by \cite{wrenninge2018synscapes}. Meanwhile, the ACDC dataset by \cite{sakaridis2021acdc} provides semantic pixel-level annotations for 19 cityscapes classes in an adverse environment. To simulate fog particles in any LiDAR-based dataset, a simulation technique by using parameters based on the See Through Fog dataset \cite{bijelic2020seeing} was proposed in \cite{hahner2021fog}. Recently, an extension to the simulation technique based on attenuation and backscattering coefficients was proposed in \cite{liu2022parallel}. 

From the literature, it becomes imperative that fog significantly affects the optical spectrum, and the particle size distribution in LiDAR-based sensors for 3D object detection plays a major role in analyzing the overall system analysis. Developing robust sensor systems requires understanding these impacts and designing for effective operation in adverse weather like fog. Therefore, in this manuscript, we are considering the role of generalized particle size distribution to first observe their effect on the overall system performance and to provide a platform for practical systems development. 
The main contributions of the manuscript are as follows:
\begin{itemize}
    \item To better understand the practical environmental conditions we created a modified version of the KITTI dataset by simulating fog particles under different fog particle size distributions and strong and moderate advection fog conditions. The created dataset is subsequently utilized in the learning of deep neural networks, which are afterwards used to predict the 3D bounding boxes.
    \item Unlike \cite{hahner2021fog}, where the backscattering coefficient is evaluated via the meteorological optical range (MOR), we evaluated the backscattering coefficient by utilizing generalized particle size distributions. First, we employed a four-parameter gamma distribution for the strong and moderate fog conditions. Further, we also considered the Junge (Power Law) distribution in moderate fog environments for modeling the fog particle size. Both distributions result in the varied point cloud generation under different foggy conditions and help create practical models. 
    \item Based on the fog particle size distribution, the system's accuracy was validated for 200 epochs for Car, Cyclist and Pedestrian scenarios. From accuracy analysis, it is observed that although the accuracy of the Car scenario is better than that of the Cyclist and Pedestrian cases due to the dimensional size variation, the system's overall accuracy in Cyclist and Pedestrian for moderate and strong fog is better under practical fog particle size.  
\end{itemize}

The rest of the manuscript is organized as follows. Section II delves into the methodology employed for the performance evaluation of the LiDAR-based systems, and the impact of the fog particle size distribution on the attenuation and backscattering coefficient under clear and foggy environments. The experimental results with respect to the system's accuracy performance, and a detailed discussion is undertaken in Section III.
Finally, the concluding remarks, along with the future directions, are made in Section IV.

\section{Methodology}
\subsection{LiDAR in clear weather}

According to \cite{hahner2021fog}, the mathematical representation for the LiDAR sensor is given as
\begin{equation}
    \label{eq: Received Power}
    P_{R}(R)=C_{A}\int_{0}^{\frac{2R}{c}}P_{T}(t)H\left(R-\frac{ct}{2}\right)dt,
\end{equation}
where $P_{R}(R)$ is the received power, which is a function of range R, $P_{T}(t)$ is the time-dependent transmitted signal power, $c$ is the speed of light, $C_{A}$ is the sensor constant independent of time and range, and $H(.)$ represents the total impulse response of the medium. Mathematically, $P_{T}(t)$ is expressed as
\begin{equation}
    \label{eq: Transmitted Power}
    P_{T}(t)=\begin{cases}
    P_{0}\ \text{Sin}^{2}\left(\frac{\pi t}{2\tau_{H}}\right),& \text{if } 0\leq t \leq 2\tau_{H}\\
    0,              & \text{otherwise},
\end{cases}
\end{equation}
where $P_{O}$ is the maximum power of the transmitted pulse, and $\tau_{H}$ is the half-power pulse width. Simultaneously, the total impulse response $H(R)$ of the LiDAR-based system is mathematically represented by
\begin{equation}
    \label{eq:Impulses Response}
    H(R)=H_{C}(R)H_{T}(R), 
\end{equation}
where $H_{T}(R)$ is the distance-based channel impulse response of the target, which is given as and $H_{C}(R)$ is the impulse response of the optical channel, which is expressed as 
\begin{equation}
    \label{eq: HC}
    H_{C}(R)=\frac{T^{2}(R)}{R^{2}}\xi(R). 
\end{equation}
In (\ref{eq: HC}), $T(R)$ stands
for the total one-way transmission loss and $\xi(R)$ denotes the crossover function defining the ratio of the
area illuminated by the transmitter and the part observed by the receiver of the LiDAR sensor. Mathematically, the expression for $T(R)$ is
\begin{equation}
    \label{eq: TR}
    T(R)=\text{exp}\left(-\int_{r=0}^{R}\alpha(r)dr\right).
\end{equation}
 The attenuation coefficient ($\alpha$), an intrinsic property of fog-based LiDAR systems, is considered uniform over the whole propagation range for homogenous conditions. Therefore, the expression in (\ref{eq: TR}) becomes 
\begin{equation}
\label{eq: Final TR}
    T(R)=\text{exp}(-\alpha R).
\end{equation}
Particularly for clear weather, the value of $H_{T}$ is given as 
\begin{equation}
\label{eq: HT clear weather}
    H_{T}(R) = H^{Hard}_{T}(R) = \beta_{0}\delta\left(R - R_{0}\right),
\end{equation}
where $\beta_0$ is the differential reflectivity of the target and $\delta(.)$ is the Kronecker delta function. For clear weather, the value of $H_{c}$ is modeled by
\begin{equation}
\label{eq: HC clear weather}
    H_{C}(R) = \frac{\xi(R)}{R^{2}}.
\end{equation}

By substituting (\ref{eq: HC clear weather}) and (\ref{eq: HT clear weather}) in (\ref{eq: Received Power}), the received power for the clear weather at the receiver is given as
\begin{IEEEeqnarray}{rCl}
    \label{eq: PR clear weather}
    P_{R,Cl}(R) &=& C_{A}\int_{0}^{2\tau_{H}}P_{0}\text{Sin}^{2}\left(\frac{\pi t}{2\tau_{H}}\right)\frac{\beta_{0}\delta\left(R-R_{0}-\frac{ct}{2}\right)}{R_{0}^{2}}dt.\IEEEnonumber\\
\end{IEEEeqnarray}
Using some algebraic and integral properties, the closed-form expressions for the received power are represented by
\begin{equation}
\label{eq: p_r_clear}
    P_{R,Cl}(R)=\begin{cases}
    \frac{C_{A}P_{0}\beta_{0}}{R_{0}^{2}}\\ \times\ \text{Sin}^{2}\left(\frac{\pi\left(R - R_{0}\right)}{c\tau_{H}}\right),&\text{if } R_{0} \leq R \leq R_{0} + c\tau_{H}\\
    0,              & \text{otherwise}.
\end{cases}
\end{equation}
\subsection{LiDAR in foggy weather}
Primarily in foggy weather conditions, the data received by the LiDAR sensors experiences attenuation and backscattering. Mie theory is usually utilized to quantify the attenuation and backscattering effects to model the foggy conditions. In the current literature, a precise model for calculating attenuation and backscattering coefficients is unavailable; therefore, Mie theory is employed. Simultaneously, 
to model the fog particle size distribution, Gamma and power law (Junge) distributions are usually employed \cite{awan2011prediction,petty2011modified,tampieri1976size,liu2020fog}.

The channel impulse response for the fog particles is bifurcated into soft and hard targets to model the fog scenario in the channel practically. Mathematically, the updated channel impulse response is given as
\begin{equation}
\label{eq: HT total}
    H_{T}(R) = H_{T}^{Soft}(R) + H_{T}^{Hard}(R),
\end{equation}
where $H_{T}^{Hard}$ is given by (\ref{eq: HT clear weather}). For foggy weather conditions the impulse response, $H_{T}^{Soft}$, for the fog particles is modeled as 
\begin{equation}
    \label{eq: HT of soft particle}
    H_{T}^{Soft}(R)=\beta U\left(R_{0}-R\right).
\end{equation}
where $U(.)$ is the Heaviside function. Combining (\ref{eq: HC}), (\ref{eq: Final TR}), (\ref{eq: HT clear weather}) and (\ref{eq: HT of soft particle}) the total channel impulse response in (\ref{eq:Impulses Response}) is obtained as 
\begin{equation}
    \label{eq: Total impulse response}
    H\left(R\right) = \frac{\text{exp}\left(-2\alpha R \right)\xi(R)}{R^{2}}\left(\beta U(R_{0} - R) + \beta_{0}\delta(R - R_{0})\right).
\end{equation}

Concurrently, the total received power obtained by the summation of the powers received back from both the hard and soft target is represented as 
\begin{IEEEeqnarray}{rCl}
\label{eq: Power hard}
    P_{R,fog}^{Hard}(R)&=&C_{A}\frac{\text{exp}(-2\alpha R_{O})}{R_{O}^{2}}\int_{0}^{2\tau_{H}}P_{0}\text{Sin}^{2}\left(\frac{\pi t}{2\tau_{H}}\right)\IEEEnonumber\\   &&\times\beta_{0}\ \delta\left(R - \frac{ct}{2} - R_{0}\right)dt\IEEEnonumber \\ 
    &=&\text{exp}\left(-2 \alpha R\right)P_{R,Cl}(R).
\end{IEEEeqnarray}
Substituting (\ref{eq: p_r_clear}) in (\ref{eq: Power hard}) we obtained a closed form expression for $P_{R,fog}^{Hard}(R)$. Moreover, for fog droplet conditions, the received power is mathematically modeled as 
\begin{IEEEeqnarray}{rCl}
\label{eq: p_r_soft power}
    P_{R,fog}^{Soft}(R) &=& C_{A}\ P_{0}\ \beta\int_{0}^{2\tau_{H}}\text{Sin}^{2}\left(\frac{\pi t}{2\tau_{H}}\right)\xi\left(R-\frac{ct}{2}\right)\IEEEnonumber\\ &&\times\frac{\text{exp}\left(-2\alpha\left(R-\frac{ct}{2}\right)\right)}{\left(R-\frac{ct}{2}\right)^{2}} U\left(R_{0}-R + \frac{ct}{2}\right)\ dt,\IEEEnonumber\\
\end{IEEEeqnarray}
where $U$(.) denotes the Heaviside function. Due to the intractability of the integral expression in (\ref{eq: p_r_soft power}), the closed-form expressions are not attainable. Therefore, to obtain the values of the power received by the fog particles in foggy conditions, we utilise the Simpson $1/3$ rule. Combining (\ref{eq: Power hard}) and (\ref{eq: p_r_soft power}), the total received power at the LiDAR receiver is obtained as 
\begin{equation}
    \label{eq: Total received power}
    P_{R,fog}(R) = P_{R,fog}^{Hard}(R) + P_{R,fog}^{Soft}(R).
\end{equation}

Usually, the acquisition and modeling of the LiDAR sensor data are majorly dependent on environmental conditions such as fog. 
In \cite{hahner2021fog}, an approximated value of the backscattering coefficient ($\beta$) in terms of MOR, utilized for system's performance evaluation under see-through fog (STF) dataset, is given as
\begin{equation}
    \label{eq: Calculation backscattering coeffecient using 0.0046/MOR}
    \beta = \frac{0.046}{MOR}.
\end{equation}
Considering the fixed relationship between $\beta$, and the meteorological range results is a misinterpretation of the foggy weather conditions. Therefore, to overcome this problem, we have employed Mie Theory, which can be used to calculate the attenuation and backscattering coefficients based on the fog particle size distributions.
In the next section, we will discuss the role of fog particle size distribution in modeling the attenuation and backscattering coefficients.

\subsection{Role of Fog Particle Size Distribution}
\begin{figure}
    \centering
    \includegraphics[width=1\linewidth,height=0.75\linewidth]{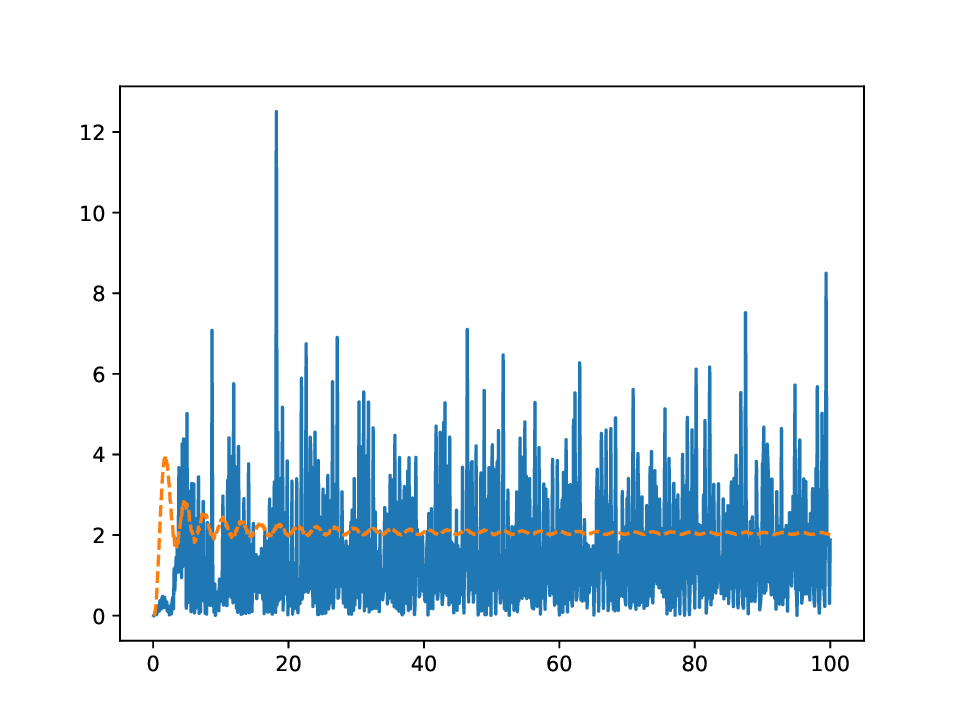}  
    \caption{Efficiency values of $Q_{ext}$ and $Q_{b}$ as a function of droplet diameter.}
    \label{fig: Qext and QB vaala}
\end{figure}

\begin{figure}
    \centering
    \includegraphics[width=1\linewidth,height=0.75\linewidth]{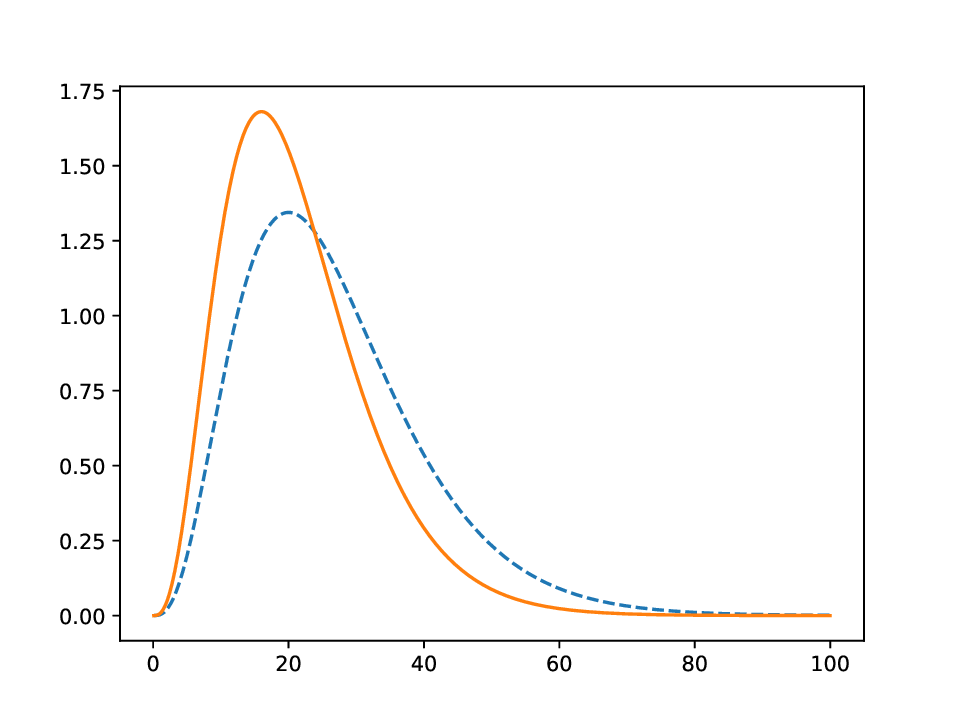}
    \caption{Effect of fog particle size distribution using Gamma distribution.}
    \label{fig: Gamma Particle Size Distribution Graph}
\end{figure}

 The scattering of laser beams from the water droplet's surface can be modeled by assuming the water droplet is a liquid sphere. Based on this assumption, the extinction efficiency, $Q_{ext}(D)$, and the backscattering efficiency, $Q_{b}(D)$, both of which are a function of the diameter of fog droplets can be calculated using Mie Theory.  Throughout the manuscript, we have considered an optical transmitting source with a wavelength of 905 nanometers (nm).
The value of both the attenuation and backscattering coefficients are shown in Fig. \ref{fig: Qext and QB vaala}. The size distribution of atmospheric water droplets is essential for attenuation coefficient calculation. Subsequently, various distributions, such as Gamma, Junge (Power-Law), etc., can be used to model the fog particle size distribution. 
Assuming non-elastic scattering, the attenuation ($\alpha$) and the backscattering ($\beta$) coefficients in terms of the fog particle size distribution are mathematically represented as
\begin{equation}
    \label{eq: Attenuation Coeffecient Calculation}
    \alpha = \frac{\pi}{8} \int_{D=0}^{\infty} D^{2}Q_{ext}(D)N(D)dD,
\end{equation}
\begin{equation}
    \label{eq: Backscattering Coeffecient Calculation}
    \beta = \frac{\pi}{8} \int_{D=0}^{\infty} D^{2}Q_{b}(D)N(D)dD,
\end{equation}
where $N(D)$ is the probability of hitting a droplet of diameter $D$. The water droplet's statistical distribution featuring different $D$ is utilized to model the foggy conditions.  Using a four-parameter ($a,\rho,\gamma, r_C$) gamma distribution, the mathematical representation of $N(D)$ is given as \cite{rasshofer2011} 
\begin{equation}
    \label{Gamma Distribution}
    N(D) = \frac{\gamma\rho b^{\frac{a+1}{\gamma}}}{\Gamma \left(\frac{a+1}{\gamma}\right)}\left(\frac{D}{2}\right)^{a}\text{exp}\left(-b\left(\frac{D}{2}\right)^{\gamma}\right),
\end{equation}
where $\Gamma(.)$ denotes gamma function, and parameters $a$, $b$, $\gamma$, and $r_C$ are positive and real, with $a$ being an integer. The parameters depend on each other and are determined by the distribution of the experimental measurements. Mathematically, parameter $b$ is expressed as 
\begin{equation}
    \label{Calculation of b using gamma distribution}
        b = \frac{a}{\gamma\left(\frac{D_{c}}{2}\right)^\gamma}.
\end{equation}
In (\ref{Gamma Distribution}) and (\ref{Calculation of b using gamma distribution}), $\Gamma(.)$ denotes the gamma function, and $D_{c}$ is the droplet radius size having maximum probability respectively.
Table \ref{tab: Gamma Distribution coeffecients value} represents strong and moderate advection fog parameters. Utilizing the parameters of Table \ref{tab: Gamma Distribution coeffecients value}, the Fig. \ref{fig: Gamma Particle Size Distribution Graph} shows the fog particle droplet size distribution for strong and moderate advection fog. Substituting (\ref{Gamma Distribution}) in (\ref{eq: Attenuation Coeffecient Calculation}) and (\ref{eq: Backscattering Coeffecient Calculation}), and utilizing the parameters' values of different foggy conditions from Table \ref{tab: Gamma Distribution coeffecients value}, the value of the attenuation and backscattering coefficient without approximation is obtained. Table \ref{tab: Values of attenuation and backscattering coeffecient obtained} represents the values of $\alpha$ and $\beta$ in different fog environments with different particle size distributions.
\begin{table}
\centering
\caption{Practical Parameters for Advection Fog  \cite{liu2022parallel}.}
\label{tab: Gamma Distribution coeffecients value}
\resizebox{\columnwidth}{!}{%
\begin{tabular}{|c|c|c|c|c|}
\hline
         Weather Conditions&  $\rho \left(cm^{-3}\right)$ &  a&  $\gamma$& $r_{C} \left(\mu m\right)$\\ \hline 
         Strong Advection Fog&  20&  3&  1& 10\\ \hline
         Moderate Advection Fog&  20&  3&  1& 8\\ \hline
\end{tabular}}
\end{table}

\begin{table}[h]
\label{tab:Results Table}
\centering
\caption{Values of $\alpha$ and $\beta$ in Strong and Moderate Advection Fog.}
\label{tab: Values of attenuation and backscattering coeffecient obtained}
\resizebox{\columnwidth}{!}{%
\begin{tabular}{|c|c|c|c|c|}
\hline
Weather Conditions& Distribution & $\alpha$ & $\beta$ & $\beta$ calculation\\ \hline 
         Strong Advection Fog& Gamma &0.028995 &  0.0011& Using MOR(40m)\\ \hline 
         Strong Advection Fog & Gamma & 0.028996 & 0.020243 & Using distribution\\ \hline
 Moderate Advection Fog& Gamma & 0.018727& 0.00057& Using MOR(80m)\\ \hline
 Moderate Advection Fog & Gamma & 0.018721 & 0.012894 & Using distribution\\ \hline
 Moderate Advection Fog & Junge & 0.026201 & 0.019104 & Using distribution\\ \hline
 
\end{tabular}%
}
\end{table}
Simultaneously, Junge distribution (Power-Law distribution) also models the moderate advection fog environment. Mathematically, the fog particle size distribution in moderate advection fog is given as 
\begin{equation}
    \label{eq: Junge Distribution}
    N(D) = 131.5\ \text{e}^{-1.76}.
\end{equation}
\begin{figure}
    \centering
    \includegraphics[width=1\linewidth,height=0.75\linewidth]{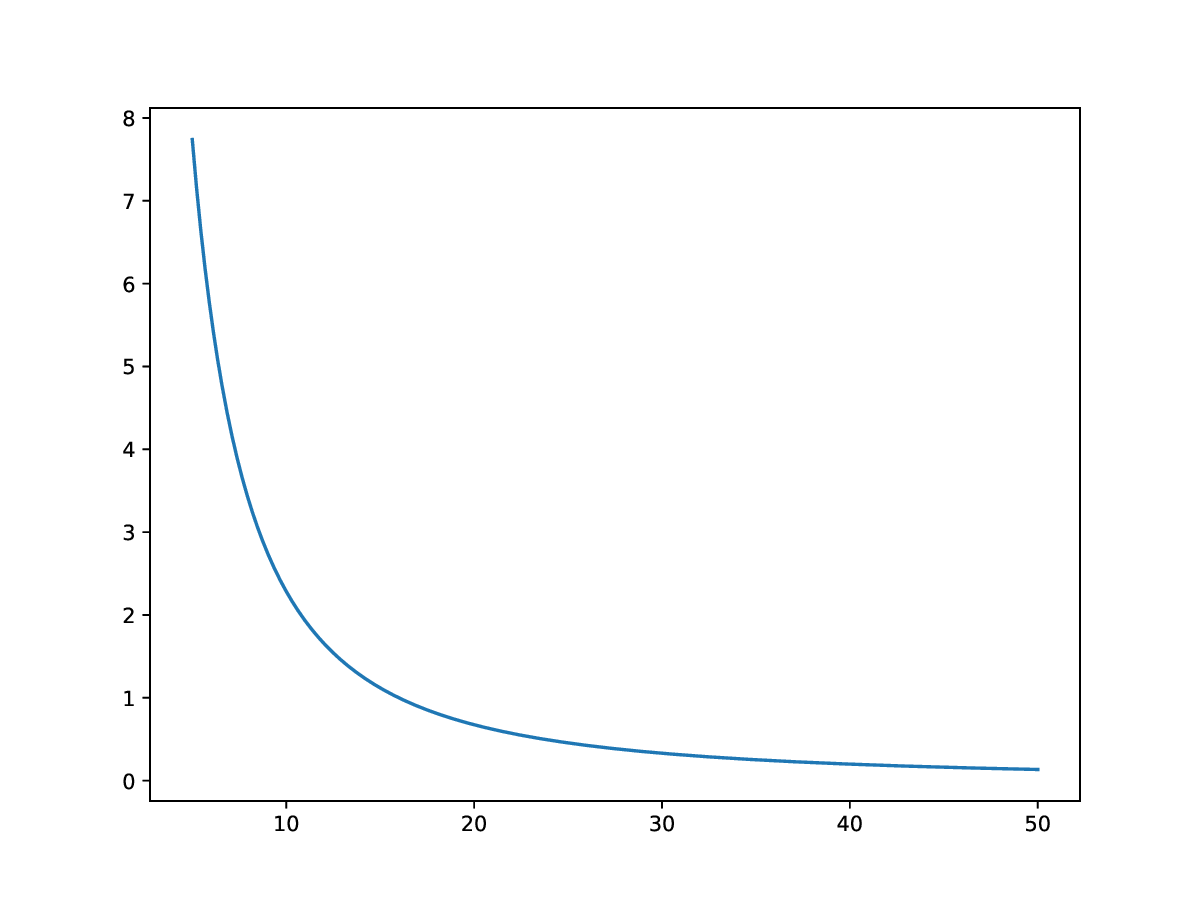}
    \caption{\footnotesize{Effect of fog particle size distribution using Junge distribution.}}
    \label{fig:Junge distribution}
\end{figure}
Fig. \ref{fig:Junge distribution}, shows the pictorial representation of the fog particle size distribution in the case of moderate advection fog. 
By substituting (\ref{eq: Junge Distribution}) in (\ref{eq: Attenuation Coeffecient Calculation}) and (\ref{eq: Backscattering Coeffecient Calculation}), the value of $\alpha$ and $\beta$ are obtained as 0.026201 and 0.019104 respectively.

\begin{figure}
     \centering
     \begin{subfigure}[b]{0.23\textwidth}
         \centering
         \includegraphics[width=\textwidth]{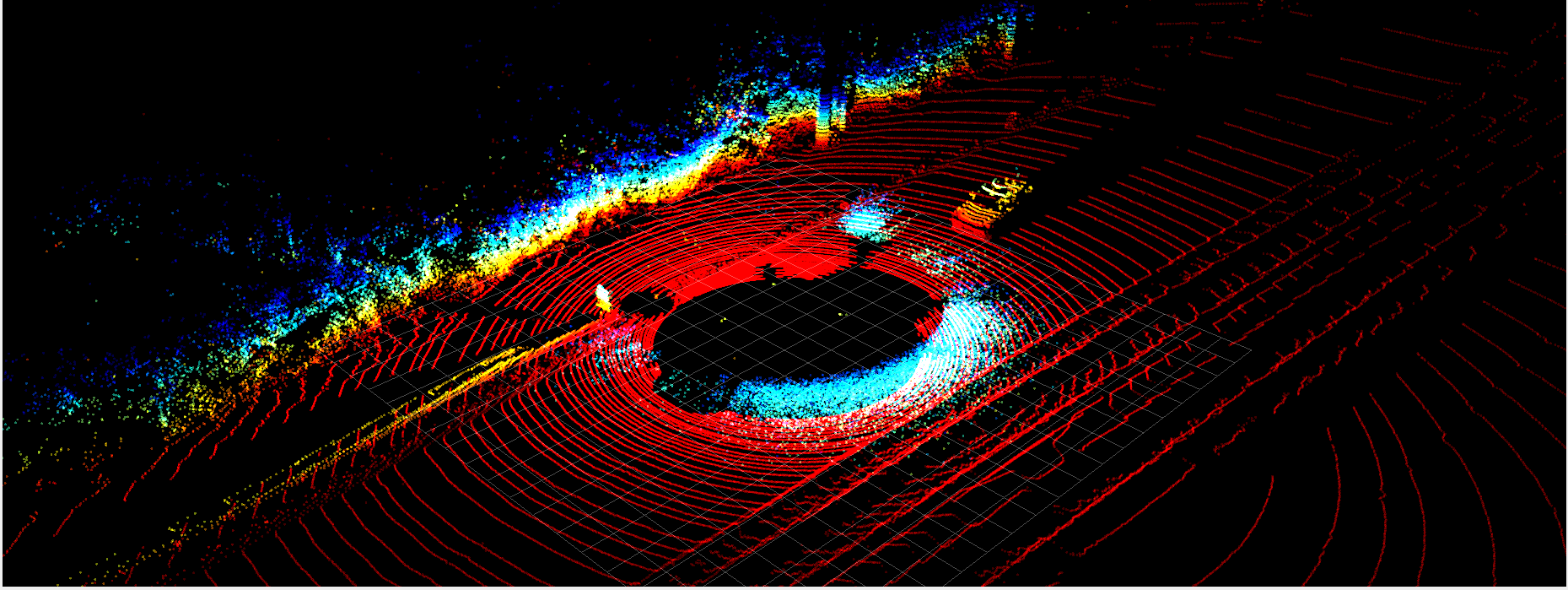}
         \caption{}
         \label{fig: Strong fog without approximation}
     \end{subfigure}
     \hfill
     \begin{subfigure}[b]{0.23\textwidth}
         \centering
         \includegraphics[width=\textwidth]{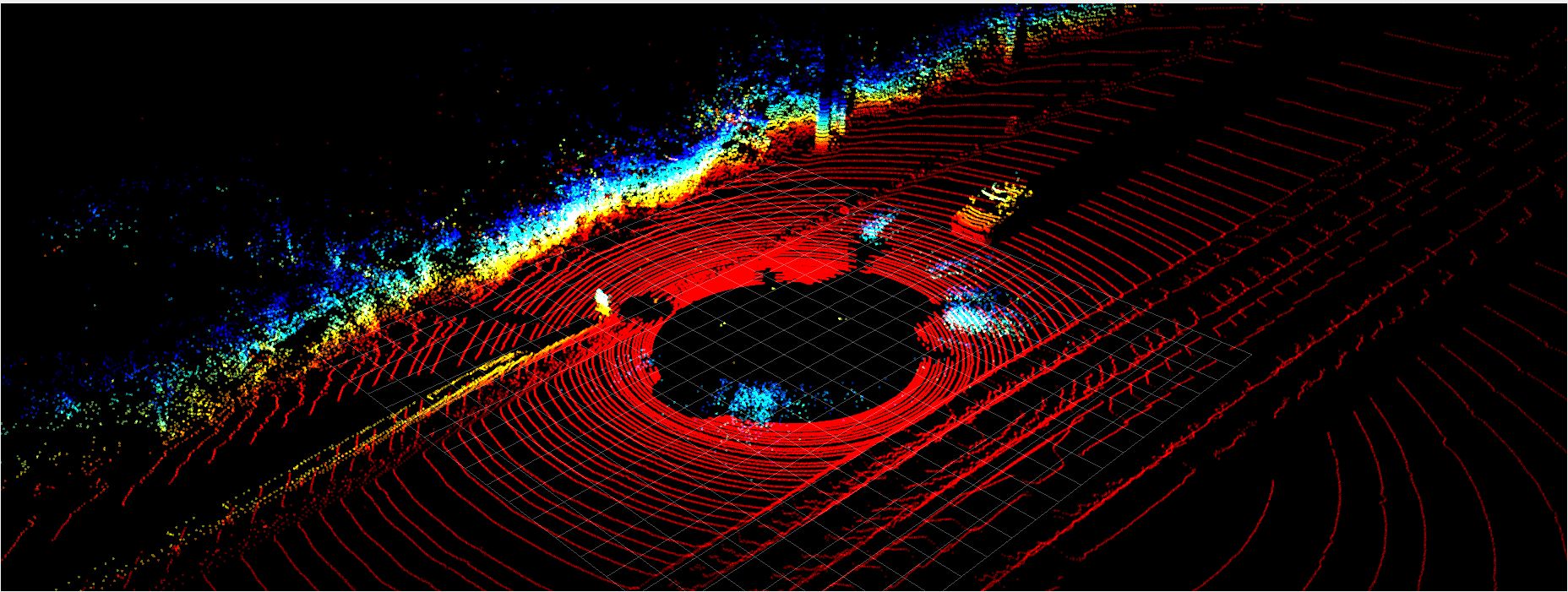}
         \caption{}
         \label{fig: Moderate advection fog without approximation}
     \end{subfigure}
     \hfill
     \begin{subfigure}[b]{0.23\textwidth}
         \centering
         \includegraphics[width=\textwidth]{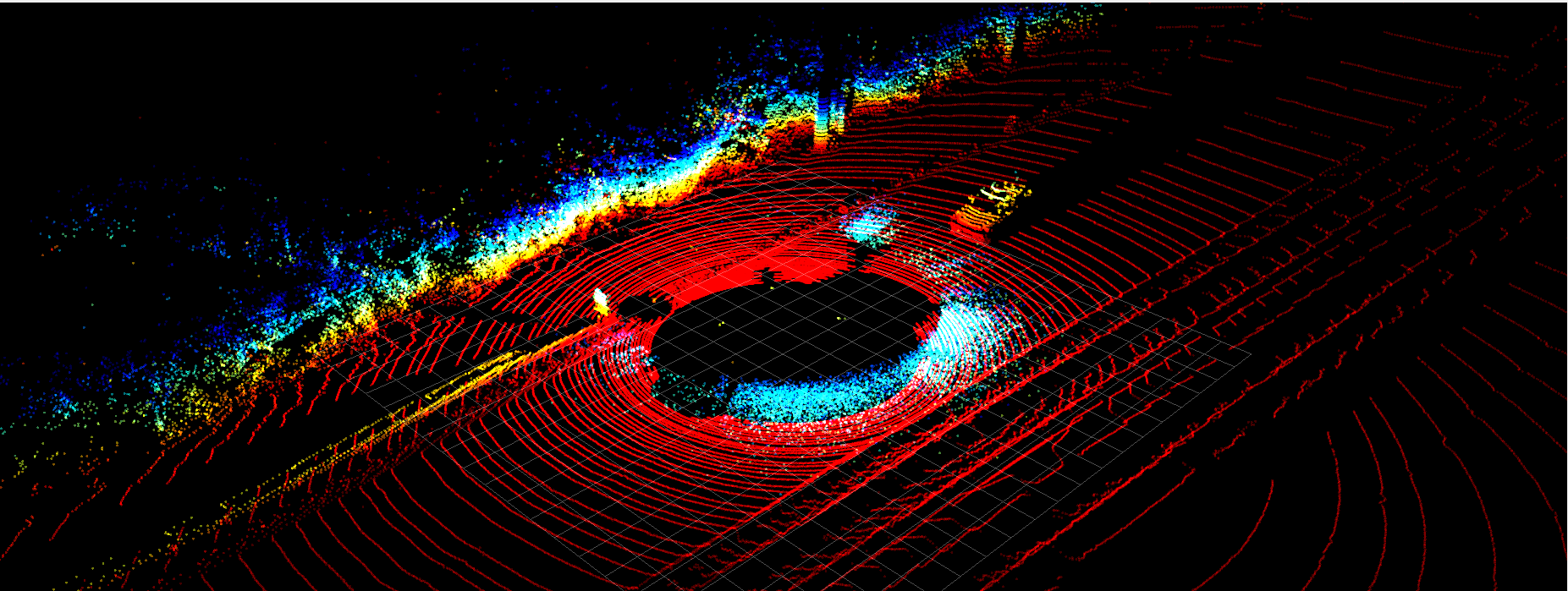}
         \caption{}
         \label{fig: Strong advection fog with approximation}
     \end{subfigure}
     \hfill
     \begin{subfigure}[b]{0.23\textwidth}
         \centering
         \includegraphics[width=\textwidth]{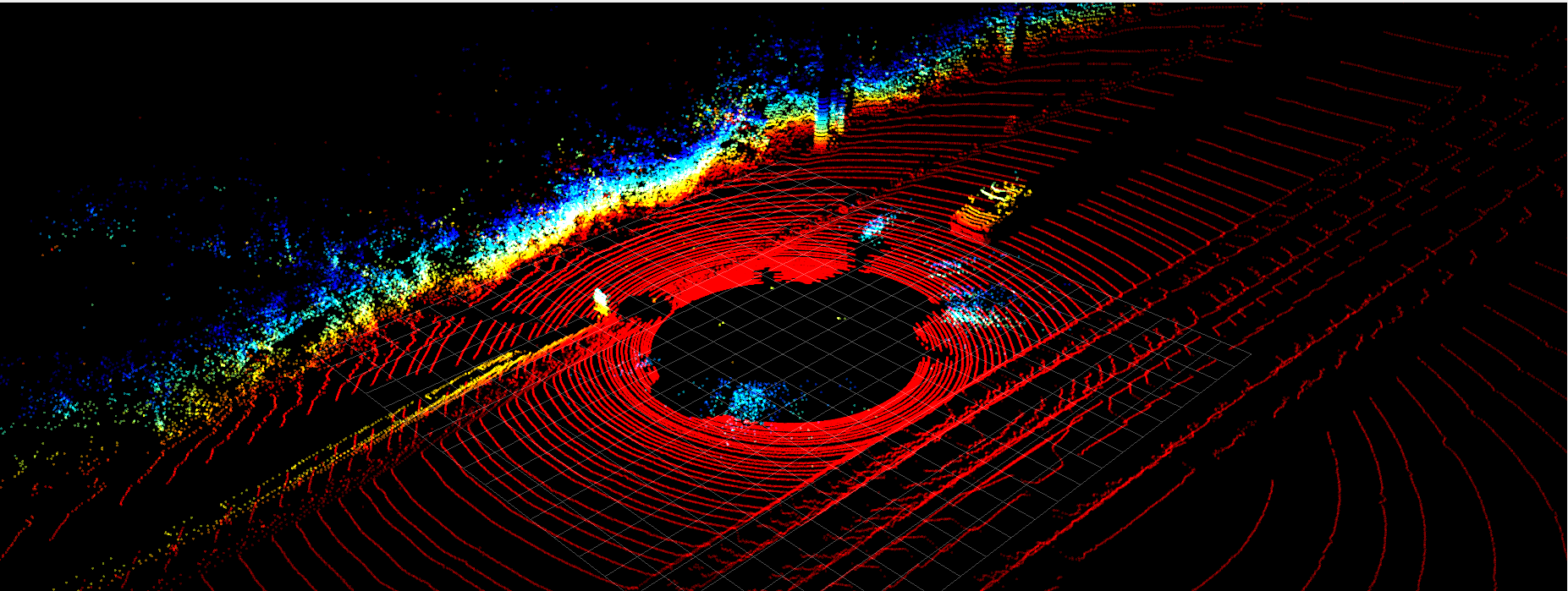}
         \caption{}
         \label{fig: Moderate advection fog with approximation}
     \end{subfigure}
    \begin{subfigure}[b]{0.23\textwidth}
         \centering
         \includegraphics[width=\textwidth]{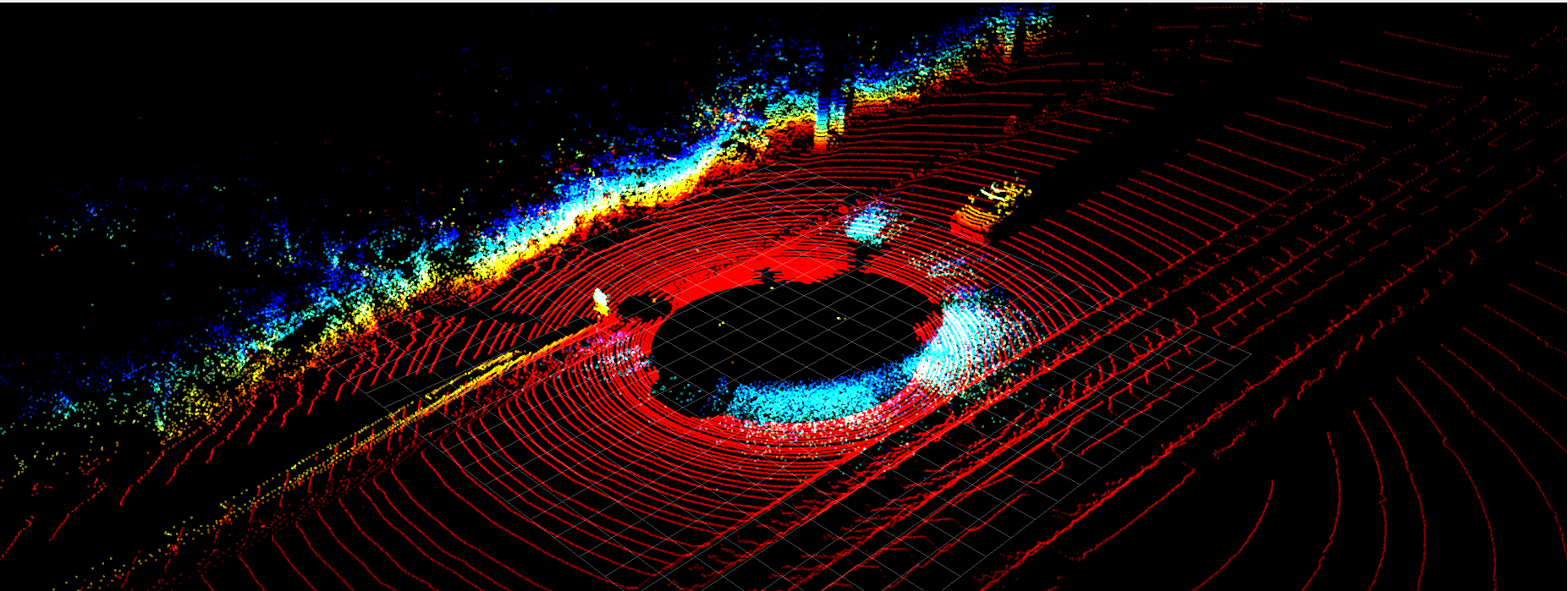}
         \caption{}
         \label{fig: Junge distributino moderate fog 131 vaala}
    \end{subfigure}
        \caption{Point clouds in different foggy environments: (4.a.) Shows strong advection fog with $\beta$ from (\ref{eq: Backscattering Coeffecient Calculation}); (4.b.) Shows moderate advection fog with $\beta$ from (\ref{eq: Backscattering Coeffecient Calculation}); (4.c.) Shows strong advection fog with $\beta$ from (\ref{eq: Calculation backscattering coeffecient using 0.0046/MOR}); (4.d.) shows moderate advection fog with $\beta$ from (\ref{eq: Calculation backscattering coeffecient using 0.0046/MOR}); (4.e.) Shows moderate advection fog with $\beta$ from (\ref{eq: Junge Distribution}).
        }
        \label{fig:Simulated Fog particles using different particle size distributions}
\end{figure}

For a detailed qualitative analysis, the point cloud scene representation of the LiDAR sensor data is shown in Fig. \ref{fig:Simulated Fog particles using different particle size distributions}. The figure shows that the point cloud representation of the LiDAR-based depends on the weather conditions. Each point in the point cloud gives a pictorial representation of the external surface of an object and usually provides information related to the position and intensity of the point. Depending on the object information, the points in point clouds are coloured accordingly. For instance, point clouds at low heights are usually highlighted in red colour, and the objects with higher heights are depicted in blue colour. Fig. \ref{fig: Strong fog without approximation} and Fig. \ref{fig: Moderate advection fog without approximation}, show the strong and moderate advection fog conditions with $\beta$ values obtained from (\ref{eq: Backscattering Coeffecient Calculation}). Simultaneously, Fig. \ref{fig: Strong advection fog with approximation} and Fig. \ref{fig: Moderate advection fog with approximation} represent the strong and moderate foggy environment based on $\beta$ obtained from (\ref{eq: Calculation backscattering coeffecient using 0.0046/MOR}). From the point cloud scene, it is observed that compared to the fog particle size distribution scenario utilizing MOR for $\beta$ value calculation and LiDAR-based sensor modeling leads to lesser point generation in the point cloud, resulting in the misinterpretation of the structural feature of the 3D environment. Therefore, for modeling the practical LiDAR-based systems the role of particle size distribution is significant.
\section{Results and Discussions}

This section, presents the results of the proposed methodology, to validate the role of the particle size distribution for Car, Cyclist and Pedestrian scenarios under strong and moderate advection fog environments. We ran the simulations for 200 epochs to validate our results and subsequently obtained the accuracy plots. Here, we considered the four-parameter Gamma distribution and Junge (Power-Law) distribution for strong and moderate fog conditions to model the fog droplet size. The KITTI data set is utilized for data generation in practical weather environments. We employed the PV-RCNN++ model to train the enhanced dataset, which is a modified version of the existing KITTI dataset.
\subsection{Strong Advection Fog Environment Analysis Using Fog Particle Size Distribution (Gamma Distribution)}
\begin{figure}[!h]
     \centering
     \begin{subfigure}[b]{0.23\textwidth}
         \centering
         \includegraphics[width=\textwidth]{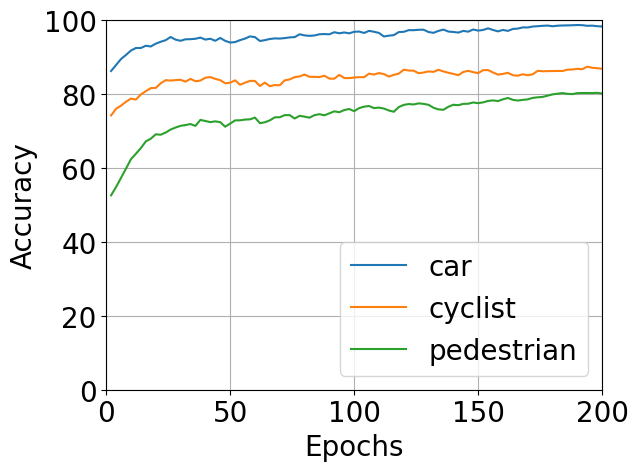}
         \caption{}
         \label{figure_3D_5a}
     \end{subfigure}
     \hfill
     \begin{subfigure}[b]{0.23\textwidth}
         \centering
         \includegraphics[width=\textwidth]{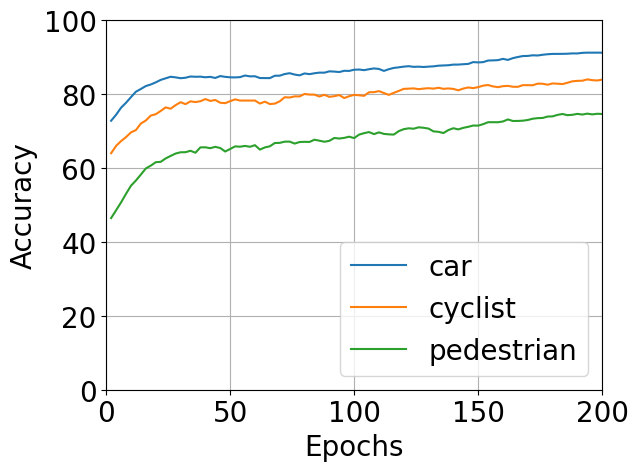}
         \caption{}
         \label{figure_3D_5b}
     \end{subfigure}
     \hfill
     \begin{subfigure}[b]{0.23\textwidth}
         \centering
         \includegraphics[width=\textwidth]{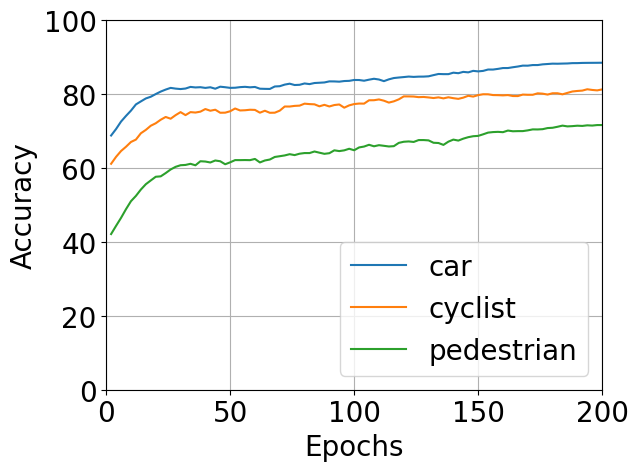}
         \caption{}
         \label{figure_3D_5c}
     \end{subfigure}
        \caption{Accuracy versus epochs for 3D bounding boxes under strong fog conditions using Gamma distribution.}
        \label{fig: Accuracy vs epochs for 3D bounding boxes in case of strong fog conditions using gamma distribution without approximation for the value of attenuation coeffecient.}
\end{figure}

Fig. \ref{figure_3D_5a}, Fig. \ref{figure_3D_5b}, and Fig. \ref{figure_3D_5c} present the accuracy versus epochs graph for 3D bounding boxes under strong advection fog for Car, Cyclist and Pedestrian cases in easy, medium, and hard conditions, respectively. Based on the plots, it is observed that the system's performance concerning the system's accuracy shows an increasing trend and then saturates with increasing epochs. Compared to medium and hard conditions the accuracy for easy conditions is higher. This can be attributed to the fact that because of the presence of other objects in the optical signal path, the number of points representing point clouds of the car decreases, resulting in an overall decrease in the system's performance. Moreover, for all three conditions (easy, medium, and hard), the system's accuracy performance is the best for the Car scenario. Particularly, the system achieved more than 98\% accuracy in easy conditions compared to 90\% and 88\% accuracy in medium and hard conditions. This is due to the reduction in the object's dimensionality that needs to be detected.  
\begin{figure}[!h]
     \centering
     \begin{subfigure}[b]{0.23\textwidth}
         \centering
         \includegraphics[width=\textwidth]{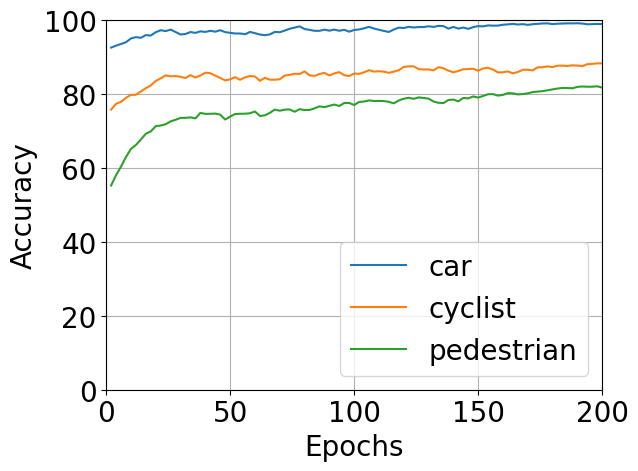}
         \caption{}
         \label{figure_BEV_6a}
     \end{subfigure}
     \hfill
     \begin{subfigure}[b]{0.23\textwidth}
         \centering
         \includegraphics[width=\textwidth]{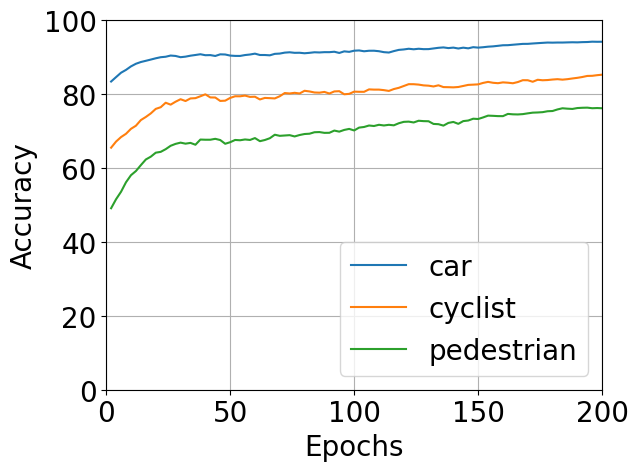}
         \caption{}
         \label{figure_BEV_6b}
     \end{subfigure}
     \hfill
     \begin{subfigure}[b]{0.23\textwidth}
         \centering
         \includegraphics[width=\textwidth]{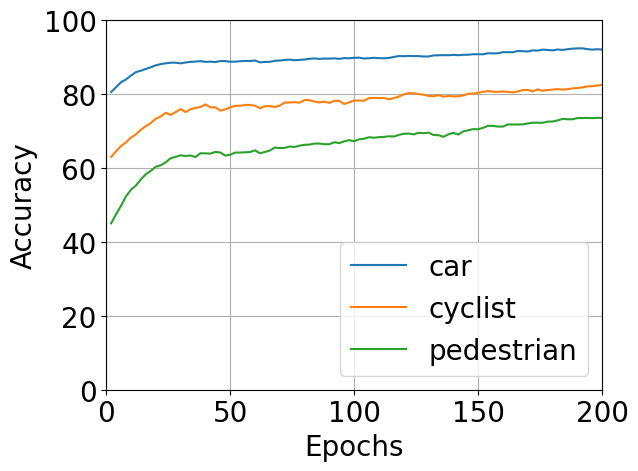}
         \caption{}
         \label{figure_BEV_6c}
     \end{subfigure}
        \caption{Accuracy versus epochs for BEV bounding boxes under strong fog conditions using Gamma distribution.}
        \label{fig: Accuracy vs epochs for BEV bounding boxes in case of strong fog conditions using gamma distribution without approximation for the value of attenuation coeffecient.}
\end{figure}

Fig. \ref{figure_BEV_6a}, Fig. \ref{figure_BEV_6b}, and Fig. \ref{figure_BEV_6c} shows the accuracy versus epochs graph of the system's performance on the birds-eye-view map of the LiDAR scene under strong advection fog in easy, moderate, and strong difficulties respectively. From the figure, it was observed that the system's performance improves with the object's increasing dimensionality. This is prevalent 
because increasing dimensionality significantly improves accuracy, with accuracy being highest for the Car and lowest for the Pedestrian. Simultaneously, an increase in the difficulty condition for object detection adversely impacts the overall accuracy with a significant decrease in the accuracy of the Car scenario with increasing 3D object detection difficulty from easy to medium and to hard. However, for the Cyclist and Pedestrian scenarios, the deterioration in the accuracy is not significant. Particularly, for the bird-eye-view detection in hard conditions, the proposed system achieves an accuracy of nearly 77\% for the Pedestrian case. This is primarily due to the practical modeling of the Fog particle size distribution in terms of a four-parameter Gamma distribution. Fog particle size distribution modeling provides improved accuracy with consideration regarding the physical traits of the environment.

\subsection{Moderate Advection Fog Environment Analysis Using Fog Particle Size Distribution}

\begin{figure}[!h]
     \centering
     \begin{subfigure}[b]{0.23\textwidth}
         \centering
         \includegraphics[width=\textwidth]{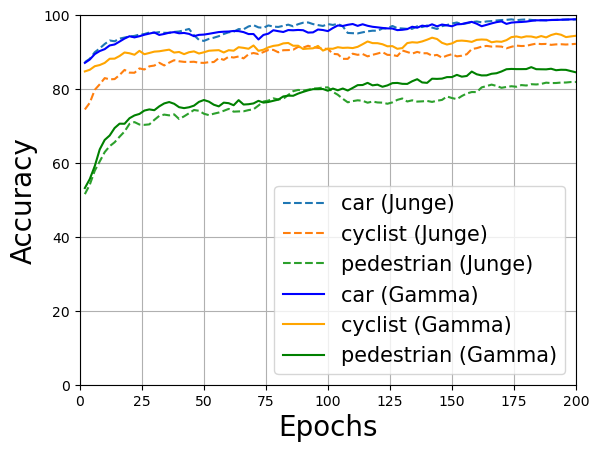}
         \caption{}
         \label{figure_7a}
       
     \end{subfigure}
     \hfill
     \begin{subfigure}[b]{0.23\textwidth}
         \centering
         \includegraphics[width=\textwidth]{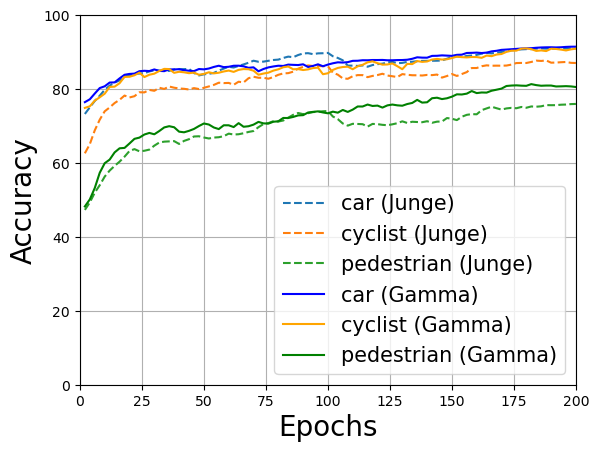}
         \caption{}
         \label{figure_7b}
     \end{subfigure}
     \hfill
     \begin{subfigure}[b]{0.23\textwidth}
         \centering
         \includegraphics[width=\textwidth]{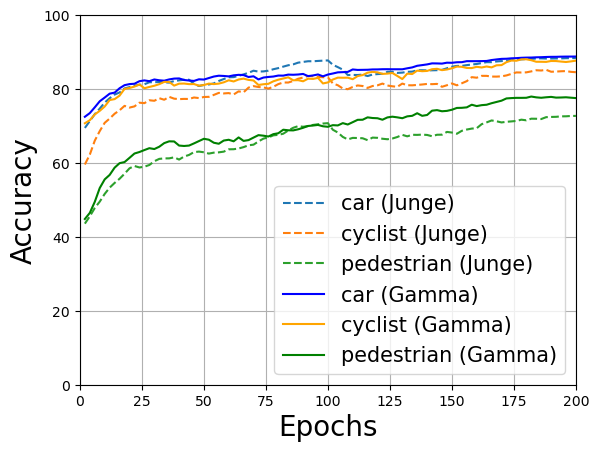}
         \caption{}
         \label{figure_7c}
     \end{subfigure}
        \caption{Accuracy versus epoch graph for 3D bounding boxes under moderate fog conditions using fog particle size distribution. Solid lines correspond to the Gamma distribution, and dotted lines correspond to Junge distribution.}
        \label{fig: Accuracy vs epochs graph of gamma distribution for moderate fog without approximation for beta}
\end{figure}



In Fig. \ref{figure_7a}, Fig. \ref{figure_7b}, and Fig. \ref{figure_7c}, the accuracy versus epochs values of the 3D bounding boxes, particularly for the easy, moderate and hard difficulty conditions under moderate advection fog environment and fog particle size distribution following Gamma and Junge distributions is simulated respectively. From the figures, it is observed that compared to medium and hard conditions, the overall accuracy of Car, Cyclist and Pedestrian scenarios under easy conditions is on the higher side, indicating better overall performance of the LiDAR-based sensors. Further, in medium and hard conditions, the system achieves an almost similar level of accuracy to that of Car and Cyclist case scenarios. This is primarily due to an increase in the number of obstacles between the LiDAR-based sensor and the object under detection. Moreover, the system's accuracy depends on the object's overall dimensionality, making detection of smaller objects difficult. This can be corroborated by decreased overall system accuracy as we transit our analysis from Car to Cyclist to Pedestrian. In addition, a decrease in the object's size results in the sparse nature of the point clouds, making object detection harder. Moreover, the overall dimensionality of the object also significantly impacts the system's accuracy with changes in the Fog particle size distribution. As highlighted in the plots, the system's accuracy reduces significantly in Cyclist and Pedestrian scenarios when the Junge distribution is utilized for fog particle size distribution modeling. However, in the Car scenario, the Junge distribution for fog particle size modeling is consistent with that of the Gamma distribution-based model. Therefore, there is always a comparable tradeoff between the utilization of particle size distribution and object dimensionality. 

\begin{figure}[!h]
     \centering
     \begin{subfigure}[b]{0.23\textwidth}
         \centering
         \includegraphics[width=\textwidth]{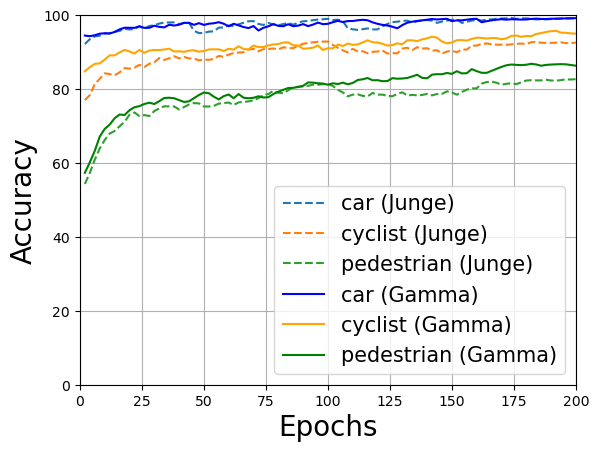}
         \caption{}
         \label{fig_8a}
         
     \end{subfigure}
     \hfill
     \begin{subfigure}[b]{0.23\textwidth}
         \centering
         \includegraphics[width=\textwidth]{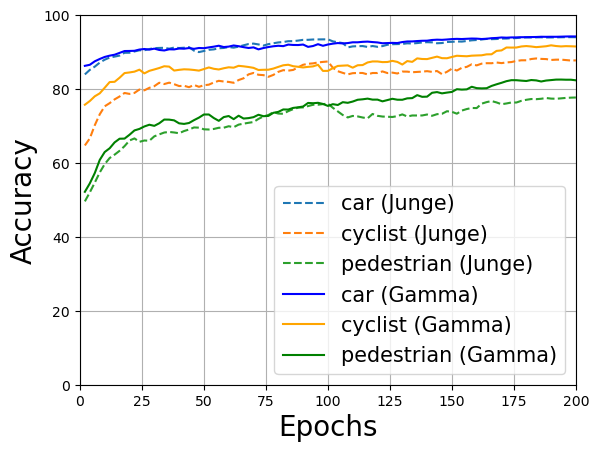}
         \caption{}
         \label{fig_8b}
     \end{subfigure}
     \hfill
     \begin{subfigure}[b]{0.23\textwidth}
         \centering
         \includegraphics[width=\textwidth]{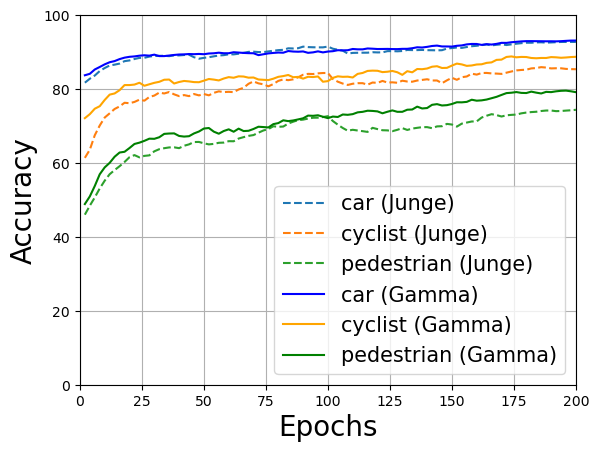}
         \caption{}
         \label{fig_8c}
     \end{subfigure}
        \caption{Accuracy versus epoch graph for BEV under moderate fog conditions using fog particle size distribution. Solid lines correspond to the Gamma distribution, and dotted lines correspond to Junge distribution.}
        \label{fig:Car BEV on moderate gamma without approximation.}
\end{figure}
The system's accuracy on a bird-eye-view (BEV) of the LiDAR scenes under moderate advection foggy environment in easy, moderate, and hard conditions is illustrated in Fig. \ref{fig_8a}, Fig. \ref{fig_8b}, and Fig. \ref{fig_8c} respectively. The plots show that the system achieves high accuracy in the Car scenario under easy conditions. Compared to 3D bounding boxes prediction, the system's performance in the case of BEV is comparatively less affected by the moderate and hard conditions. Therefore, there is clear demarcation and no overlapping in the Car and Cyclist accuracy for moderate and hard conditions. Similar to the 3D bounding boxes prediction under a moderate fog environment, the accuracy of the car scenario is also high in the BEV case. This is majorly accredited to the fact that the accuracy analysis obtained from the fog particle size distribution provides robust versatility in catering for the practical implications of the LiDAR-based systems under moderate fog environments. Further, the system's accuracy in the Car scenario for all the difficulty levels and under the Junge distribution is relatively equivalent to that of the Gamma distribution. Therefore, under the Car scenario, either Gamma or Junge distribution can be utilized to model the fog particle size distribution depending on the availability of the distribution parameters. However, with a reduction in the overall dimensional size of the object (from Car to Cyclist to Pedestrian), utilization of Junge distribution as a fog particle size distribution model may lead to missed estimations and a reduction in the overall system's accuracy.

\subsection{Strong Advection Fog Environment Analysis Using MOR}

\begin{figure}[!h]
     \centering
     \begin{subfigure}[b]{0.23\textwidth}
         \centering
         \includegraphics[width=\textwidth]{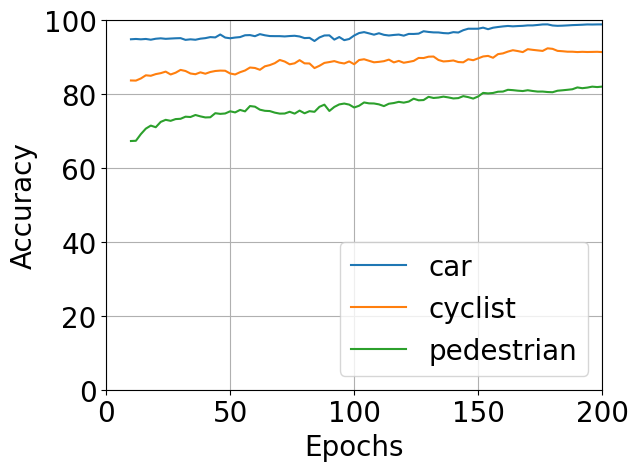}
         \caption{}
         \label{fig_9a}
     \end{subfigure}
     \hfill
     \begin{subfigure}[b]{0.23\textwidth}
         \centering
         \includegraphics[width=\textwidth]{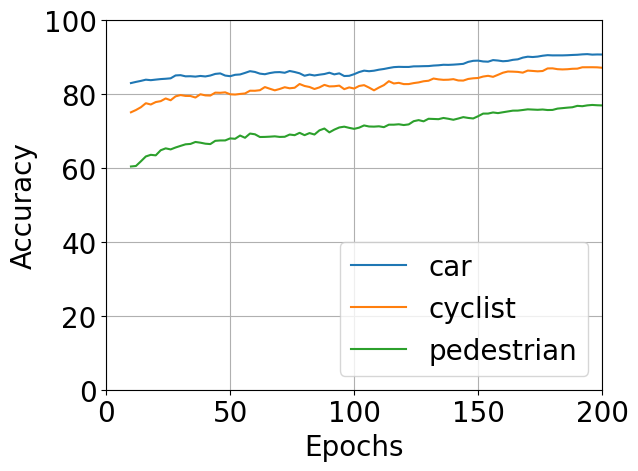}
         \caption{}
         \label{fig_9b}
     \end{subfigure}
     \hfill
     \begin{subfigure}[b]{0.23\textwidth}
         \centering
         \includegraphics[width=\textwidth]{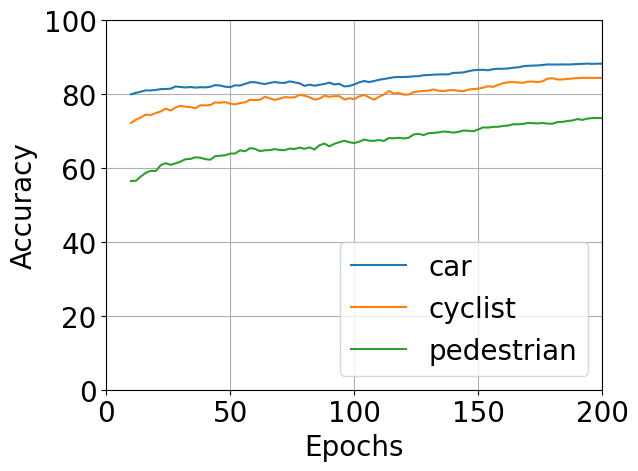}
         \caption{}
         \label{fig_9c}
     \end{subfigure}
        \caption{Accuracy versus epochs graph of 3D bounding boxes for strong fog using MOR.}
        \label{fig: Accuracy vs epochs graph of 3D bounding boxes in gamma distribution for strong fog with approximation for beta}
\end{figure}

Fig. \ref{fig_9a}, Fig. \ref{fig_9b}, and Fig. \ref{fig_9c} present the accuracy versus epochs plots for 3D bounding boxes in strong advection fog environments under easy, medium, and hard conditions while considering MOR to calculate the backscattering coefficient. From the plots, it is observed that the overall system's accuracy is better in the case of the Car, which is mainly due to the significant generation of points for point cloud due to the large dimensional feature of the Car. Since the overall system accuracy analysis is based on $\beta$ values, therefore, obtaining $\beta$ values from the MOR generates fewer points for the point cloud. Overall, This increases the system's accuracy but will lead to prediction errors in practical implementation. Moreover, the accuracy performance gap between the Car and the Cyclist decreases with increased difficulty; i.e., for easy difficulty, the Car-Cyclist gap is higher than for moderate and hard difficulty. This is due to the dimensional relationship between the object to be detected and the presence of other obstacles in the environment.

\begin{figure}[!h]
     \centering
     \begin{subfigure}[b]{0.23\textwidth}
         \centering
         \includegraphics[width=\textwidth]{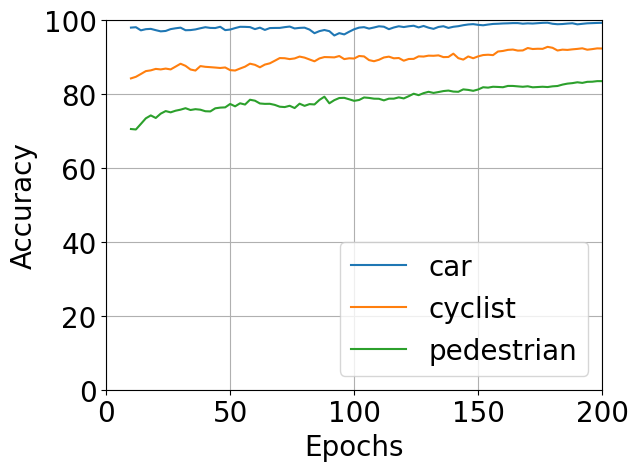}
         \caption{}
         \label{fig_10a}
     \end{subfigure}
     \hfill
     \begin{subfigure}[b]{0.23\textwidth}
         \centering
         \includegraphics[width=\textwidth]{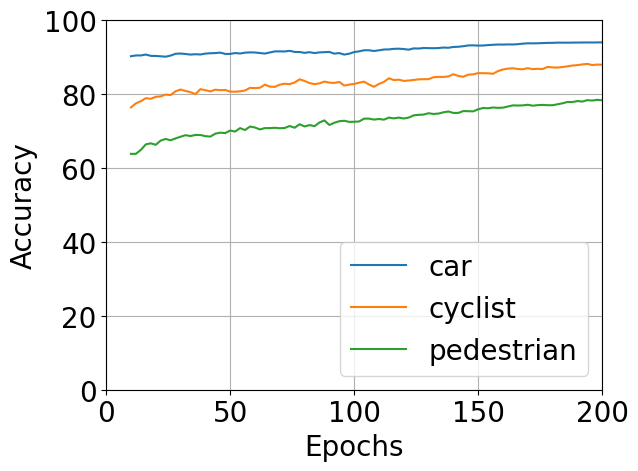}
         \caption{}
         \label{fig_10b}
     \end{subfigure}
     \hfill
     \begin{subfigure}[b]{0.23\textwidth}
         \centering
         \includegraphics[width=\textwidth]{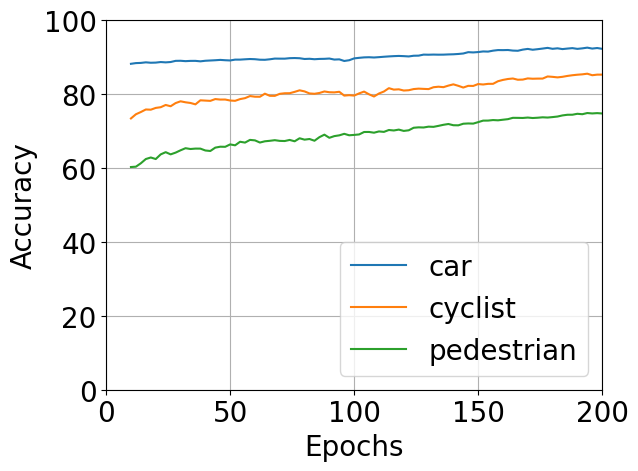}
         \caption{}
         \label{fig_10c}
     \end{subfigure}
        \caption{Accuracy versus epoch plots of BEV under strong fog using MOR.}
        \label{fig: Accuracy vs epochs graph of BEV bounding boxes in gamma distribution for strong fog with approximation for beta}
\end{figure}

The accuracy of the system on the birds-eye-view map of the LiDAR scene in a strong advection fog environment with $\beta$ values from MOR expression is shown in Fig. \ref{fig_10a}, Fig. \ref{fig_10b}, and Fig. \ref{fig_10c} respectively. From the figure, it is inferred that the system's accuracy is best in the Car scenario and worst in the Pedestrian scenario. This is primarily due to the fact that, with increasing difficulty, the overall accuracy of LiDAR-based systems decreases with the overall accuracy of the Car falling from above 98\% in easy difficulty to around 92\% and below 90\% in moderate and hard difficulties, respectively.

\begin{table*}[!h]
\centering
\caption{Average precision of PV-RCNN++ on simulated fog dataset at a given intersection over union (IOU).}
    \label{Table_3}
    \begin{tabular}{|c|c|c|c|c|c|c|c|c|c|l|l|l|l|} \hline  
         \multirow{2}{*}{Method} & \multicolumn{3}{c}{Car AP@0.7 IOU} & \multicolumn{3}{|c}{Cyclist AP@0.5 IOU} & \multicolumn{3}{|c}{Pedestrian AP@0.5 IOU} & \multicolumn{3}{|c|}{mAP over classes}\\ 
         \cline{2-13}
         &  easy & med & hard & easy & med & hard & easy & med & hard & easy & med & hard\\
         \hline
         Strong Advection Fog (using MOR) & 98.85 & 90.64 & 88.09 &  
         91.18 & 86.83 & 84.27 & 
         82.31 & 76.81 & 73.41 & 
         90.78 & 84.76 & 81.92\\
         \hline
         Strong Advection Fog & 
         98.10 & 91.17 & 88.51 & 
         86.49 & 84.40 & 81.85 & 
         79.73 & 74.45 & 71.60 & 
         88.10 & 83.34 & 80.65 \\
         \hline
         Moderate Advection Fog (using MOR) & 97.91& 91.60& 89.27 & 
         95.59& 91.55 & 87.84 & 
         85.52 & 81.82& 78.50 
         &93.01 &88.32 &85.20\\
         \hline
         \begin{tabular}[c]{@{}l@{}}Moderate Advection Fog \\ (using Gamma distribution)\end{tabular} & 98.93 & 91.47 & 88.83 & 94.80 & 91.36 & 88.16 & 83.76 & 80.17 & 77.31 & 92.50 & 87.66 & 84.77 \\ 
          \hline
         \begin{tabular}[c]{@{}l@{}}Moderate Advection Fog \\ (using Junge distribution)\end{tabular}& 
         98.92&  91.08 & 88.42 & 
         92.46 & 86.92 & 84.33 & 
         81.95 & 76.06 & 72.79 & 
         91.11 & 84.69 & 81.85 \\ 
         \hline
    \end{tabular}
\end{table*}

\subsection{Moderate Advection Fog Environment Analysis Using MOR}

\begin{figure}[!h]
     \centering
     \begin{subfigure}[b]{0.23\textwidth}
         \centering
         \includegraphics[width=\textwidth]{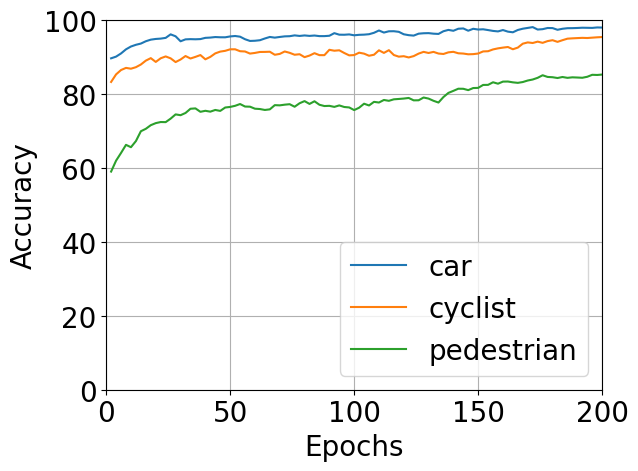}
         \caption{}
         \label{fig_11a}
     \end{subfigure}
     \hfill
     \begin{subfigure}[b]{0.23\textwidth}
         \centering
         \includegraphics[width=\textwidth]{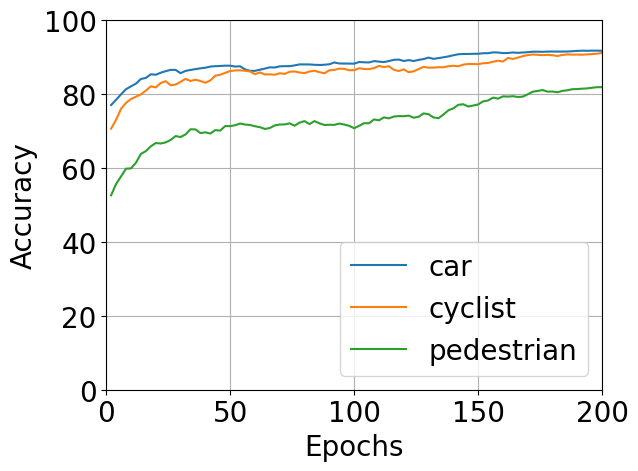}
         \caption{}
         \label{fig_11b}
     \end{subfigure}
     \hfill
     \begin{subfigure}[b]{0.23\textwidth}
         \centering
         \includegraphics[width=\textwidth]{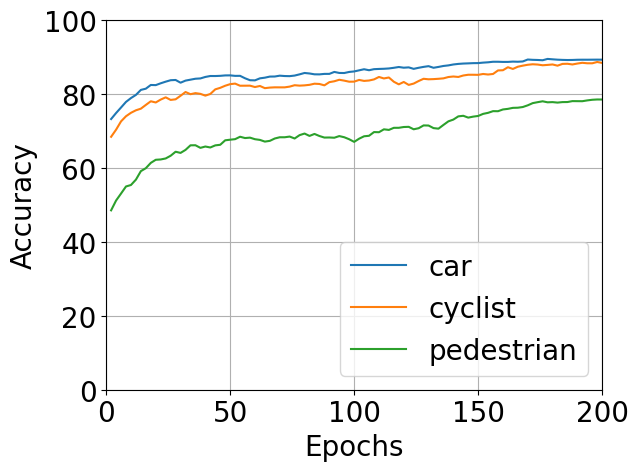}
         \caption{}
         \label{fig_11c}
     \end{subfigure}
        \caption{Accuracy versus epochs graph of 3D bounding boxes for moderate fog using MOR.}
        \label{fig: Accuracy vs epochs graph of 3D bounding boxes in gamma distribution for moderate fog with approximation for beta}
\end{figure}

Fig. \ref{fig_11a}, Fig. \ref{fig_11b}, and Fig. \ref{fig_11c} show the accuracy performance of the system from 3D bounding boxes prediction in case of moderate fog environment under easy, moderate and hard difficulty for the value of $\beta$ obtained from MOR expression. As observed from the plots, the system's accuracy performance is best in the Car scenario compared to Cyclist and Pedestrian. On the contrary, in moderate and hard difficulty, the accuracy of the LiDAR-based sensors in the Car and Cyclist scenario is nearly equal with increasing epochs. On the contrary, compared to Car and Cyclist scenarios there is a significant difference in the system's accuracy in the Pedestrian scenario. This signifies the role of dimensionality and the impact of other obstructing objects present in the path of the sensor and target object. 

\begin{figure}[!h]
     \centering
     \begin{subfigure}[b]{0.23\textwidth}
         \centering
         \includegraphics[width=\textwidth]{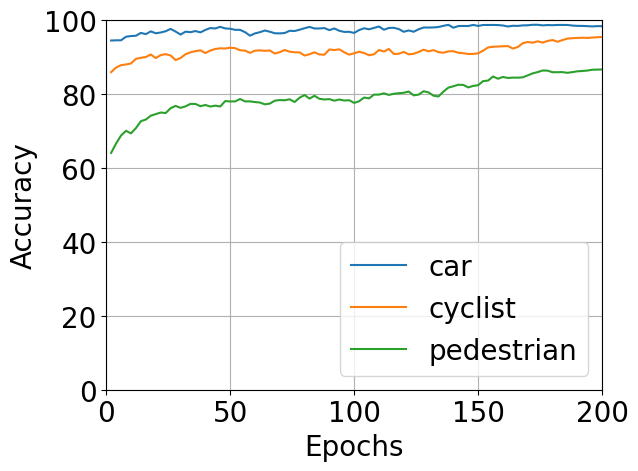}
         \caption{}
         \label{fig_12a}
     \end{subfigure}
     \hfill
     \begin{subfigure}[b]{0.23\textwidth}
         \centering
         \includegraphics[width=\textwidth]{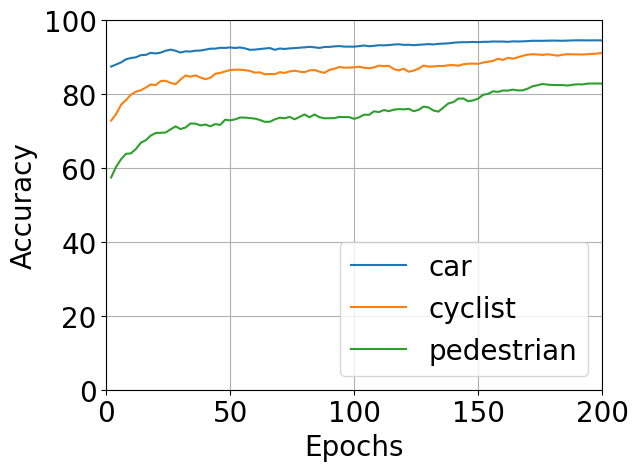}
         \caption{}
         \label{fig_12b}
     \end{subfigure}
     \hfill
     \begin{subfigure}[b]{0.23\textwidth}
         \centering
         \includegraphics[width=\textwidth]{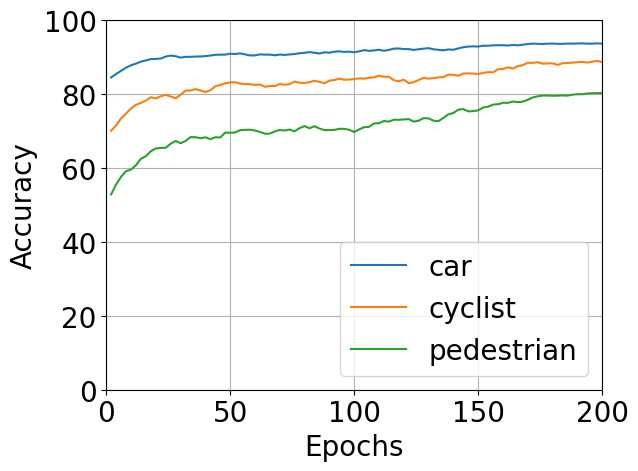}
         \caption{}
         \label{fig_12c}
     \end{subfigure}
        \caption{Accuracy versus epochs graph of BEV bounding boxes in moderate fog using MOR.}
        \label{fig: Accuracy vs epochs graph of BEV bounding boxes in gamma distribution for moderate fog with approximation for beta}
\end{figure}

Fig. \ref{fig_12a}, Fig. \ref{fig_12b}, and Fig. \ref{fig_12c} illustrate the accuracy performance of the system on birds-eye-view maps of LiDAR scenes with moderate fog simulated in easy, medium and hard difficulties with $\beta$ values obtained from MOR expression. From the plots, it is observed that the system's accuracy in the Car scenario remains stable and above 90\% across the whole range of epochs in easy conditions but shows a significant improvement for moderate and hard difficulties in Cyclists and Pedestrians scenarios. Further, the difference in performance increases in Car, Cyclist and Pedestrian scenarios due to the dimensionality variations and sparse nature of the point clouds.

Table \ref{Table_3} represents the comparative analysis of the proposed system under various fog environments and difficulties with $\beta$ values obtained from particle size distribution (Mie Theory) and MOR expression. From the table, it is observed that in strong fog conditions, the proposed system with $\beta$ obtained from fog particle size distribution performed better in the case of medium and hard car conditions in comparison to the system using MOR expression approximation for $\beta$ calculations. This can be followed from the fact that 
Further, from the mean average precision perspective, in strong fog conditions, the accuracy achieved by the system without MOR expression approximation is comparable to the accuracy achieved by the system using approximations. 
This primarily highlights the misinterpretation in terms of false prediction by the system while working with $\beta$ values obtained from the MOR expression.
The results are in concurrence with the number of fog particles simulated in the point cloud as shown in Fig. \ref{fig:Simulated Fog particles using different particle size distributions}, which indicates that the point cloud generated in case of $\beta$ obtained from Mie theory is more dense than the point cloud generated with $\beta$ obtained from MOR expression. 
Therefore, for better system performance articulation the ideal approach for modeling the effect of fog is by considering the role of Mie theory in calculation and analysis of attenuation and backscattering coefficient, which in turn are utilized for point cloud generation, and subsequent training of neural network model.

\section{Concluding Remarks} 
In this manuscript, we investigated the role of fog particle size distribution on 3D object detection for the performance evaluation of LiDAR-based systems under adverse weather conditions. 
For the system's performance evaluation, we formulated a methodology based on the Mie theory, where the attenuation and backscattering coefficients were calculated based on the fog particle size distributions and extinction and backscattering efficiency, respectively. 
Subsequently, the point clouds are generated for the strong and moderate fog environments by utilising the generalised backscattering coefficient values from Mie theory. 
We observed that the fog simulation process generated a more realistic interpretation through increased points in the point clouds under strong and moderate advection fog environments. Particularly, in a strong advection fog environment the system accuracy in Car scenarios under easy and hard conditions shows a decay of about 10\%, while for Pedestrian cases the system's accuracy reduces to 8\% under easy and hard detection difficulties.

From the mean average precision (mAP) perspective, we observed a prominent difference in the accuracy values in the case of strong fog, clearly indicating the precise role of simulating fog particles through Mie theory rather than through the generation from MOR approximation. 
However, in moderate fog conditions, the overestimation of the 3D object detection while utilizing the MOR approximation is still comparable to the values obtained from Mie's theory. 
Point cloud modeling using precise particle size distributions provides a practical realization of the system and helps attain accurate functionalities, particularly for dense fog conditions. 
The improved system's accuracy performance under simulated fog environments further corroborates the overall mathematical framework employed. 
The future work would comprise the extension of the proposed framework to incorporate more practical environments where a tradeoff between the system's efficiency and accuracy due to the presence of snow, haze and rain particles in conjunction with the fog particles.

\bibliographystyle{IEEEtran}
\bibliography{example_bib}
\end{document}